\documentclass[final]{cvpr}

\usepackage{times}
\usepackage{epsfig}
\usepackage{graphicx}
\usepackage{amsmath}
\usepackage{amssymb}
\usepackage[utf8]{inputenc}
\usepackage{algorithm, algorithmic}

\usepackage{float}
\usepackage{subfigure} 
\usepackage{multirow}
\usepackage{amsmath}
\usepackage{booktabs}       
\usepackage{stfloats}
\usepackage{wrapfig}
\DeclareMathOperator*{\argmax}{arg\,max}
\DeclareMathOperator*{\argmin}{arg\,min}

\usepackage[pagebackref=false,breaklinks=true,colorlinks,bookmarks=false]{hyperref}

\def\cvprPaperID{3193} 
\def\confYear{CVPR 2021}

\begin{document}

\title{{One-Shot Neural Ensemble Architecture Search by Diversity-Guided \\Search Space Shrinking}}


\author{Minghao Chen$^1\thanks{This work is done when Minghao is an intern at Microsoft.}$, Houwen Peng$^2\thanks{Corresponding author}$, Jianlong Fu$^2$, Haibin Ling$^1$ \\
$^1$Stony Brook University \quad $^2$Microsoft Research Asia \\
{\tt\small \{minghao.chen,haibin.ling\}@stonybrook.edu, \{hopeng,jianf\}@microsoft.com}}

\maketitle
\pagestyle{empty}
\thispagestyle{empty}

\begin{abstract}
Despite remarkable progress achieved, most neural architecture search (NAS) methods focus on searching for one single accurate and robust architecture.
To further build models with better generalization capability and performance, model ensemble is usually adopted and performs better than stand-alone models. Inspired by the merits of model ensemble, we propose to search for multiple diverse models simultaneously as an alternative way to find powerful models. Searching for ensembles is non-trivial and has two key challenges: enlarged search space and potentially more complexity for the searched model. 
In this paper, we propose a one-shot neural ensemble architecture search (NEAS) solution that addresses the two challenges. For the first challenge, we introduce a novel diversity-based metric to guide search space shrinking, considering both the potentiality and diversity of candidate operators. For the second challenge, we enable a new search dimension to learn layer sharing among different models for efficiency purposes. The experiments on ImageNet clearly demonstrate that our solution can improve the supernet's capacity of ranking ensemble architectures, and further lead to better search results. The discovered architectures achieve superior performance compared with state-of-the-arts such as MobileNetV3 and EfficientNet families under aligned settings. Moreover, we evaluate the generalization ability and robustness of our searched architecture on the COCO detection benchmark and achieve a 3.1\% improvement on AP compared with MobileNetV3. Codes and models are available \href{https://github.com/researchmm/NEAS}{here}.
\end{abstract}


\section{Introduction}
The emergence of deep neural networks greatly relieves the need for feature engineering. Previous studies have shown that the design of neural network architecture \cite{he2016deep, ma2018shufflenet, sandler2018mobilenetv2,xie2017aggregated} is essential to the performance for varied tasks in computer vision. However, the number of possible architectures is enormous, making the manual design very difficult. Neural Architecture Search (NAS) \cite{zoph2016neural} aims to automate the design process. Recently, NAS methods have achieved state-of-the-arts on varied tasks such as image classification \cite{zoph2016neural}, semantic segmentation \cite{liu2019auto}, object detection \cite{chen2019detnas}, \etc. Despite great progress achieved, most of the NAS methods focus on searching for optimal architectures of single models. However, the generalization ability and performance of single models are usually affected by different initialization, noisy data, and training recipe modification.

\begin{figure} 
\centering
\includegraphics[width=0.95\linewidth]{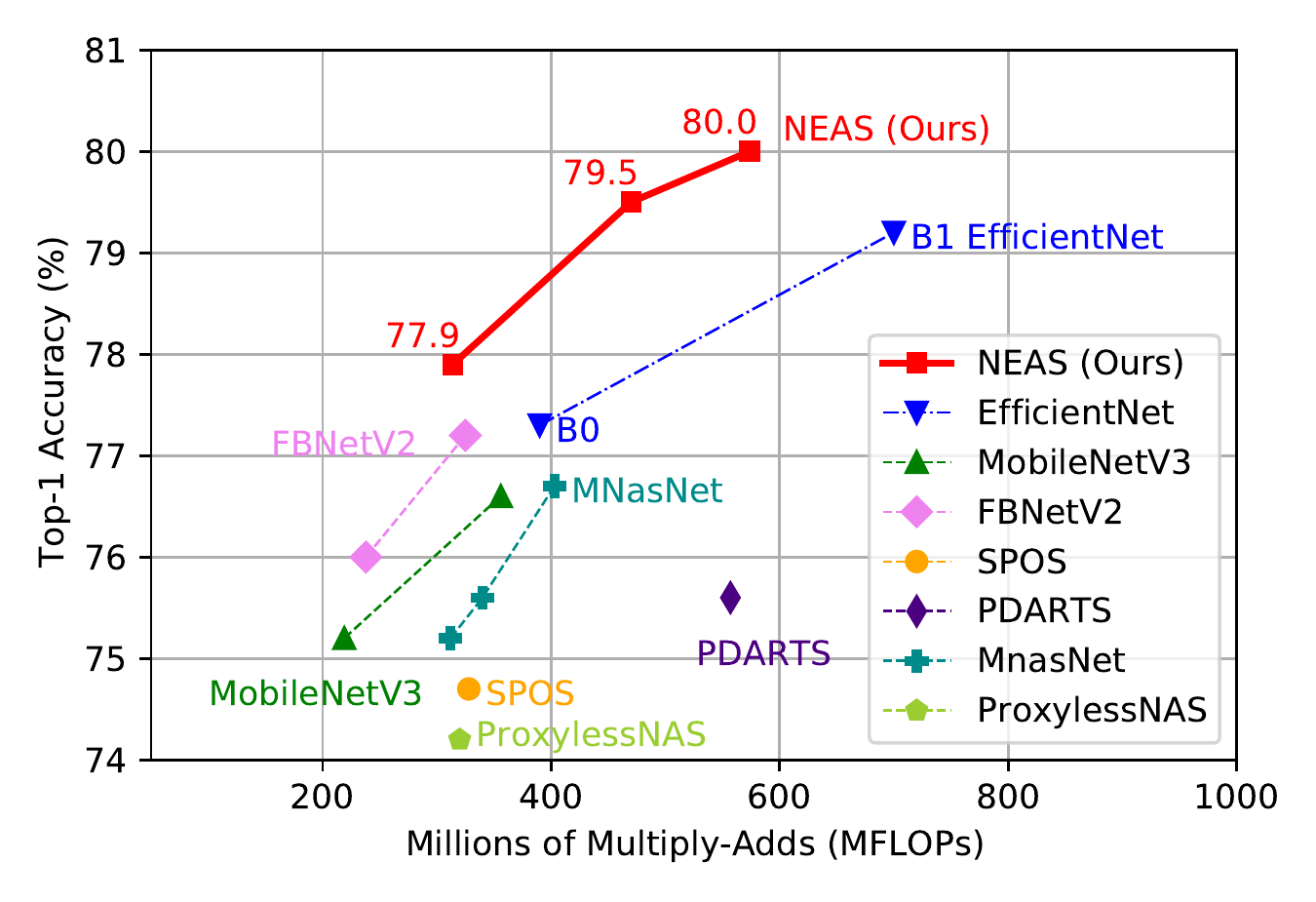} 
\caption{Comparison of our method with state-of-the-art approaches on ImageNet under mobile settings.}
\label{Fig.acc} 
\vspace{-5mm}
\end{figure}

Model ensemble has been proved to be a universally effective method to build more robust and accurate models compared with single models. Implicit ensemble methods like Dropout \cite{dropout}, Dropconnect \cite{dropconnect}, StochDepth \cite{stochasticdepth}, Shake-Shake \cite{gastaldi2017shakeshake} are already widely used in neural architecture design. On the contrary, although explicit ensemble methods like averaging, bagging, boosting, and stacking have been commonly adopted in large competitions and real-world scenarios. The use of explicit ensemble methods in designing efficient models is not fully explored due to the extra computation they brought.

Inspired by the effectiveness of ensemble, we propose to search for multiple models instead of one simultaneously to form a robust, accurate and efficient ensemble model. However, the combination of NAS and ensemble faces two challenges: (1) efficient search and supernet optimization over a large search space (2) reducing the extra complexity brought by model ensemble. Addressing these challenges, in this paper, we propose a one-shot \textit{neural ensemble architecture search} (NEAS) approach searching for lightweight ensemble models.

To solve the first challenge caused by the enlarged space of ensemble models compared with single models, we propose a novel metric called \textit{diversity score} to progressively drop inferior candidates during supernet training, thus reduce the difficulty of finding promising ensemble models. This metric explicitly quantifies the diversity between the operators, which is commonly believed to be a key factor in building models with better feature expression capability.

To solve the second challenge, we introduce the layer sharing mechanism to reduce the model complexity. We allow the ensemble components share some shallow layers and search for the best architectures of the shared layers together with the architectures of the rest layers. We further introduce a new search dimension called \textit{split point} to automatically find optimal layers for sharing under a given FLOPs constraint.

Comprehensive experiments verify the effectiveness of the proposed diversity score and layer sharing strategy. They improve the ranking ability of trained supernet and lead to better searched architectures under same complexity constraint. The searched architectures generate new state-of-the-art performance on ImageNet \cite{imagenet_cvpr09}. For instance, as shown in Fig.~\ref{Fig.acc}, our search algorithm finds a 314M FLOPs model that achieves 77.9\% top-1 accuracy on ImageNet, which is 19\% smaller and 1.6\% better than EfficientNet-B0 \cite{tan2020efficientnet}. The architecture discovered by NEAS transfers well to downstream object detection task, suggesting the generalization ability of the searched models. We obtain an AP of 33.0 on COCO validation set, which is superior to the state-of-the-art backbone, MobileNetV3 \cite{howard2019searching}. 

In summary, we make the following contributions:
\begin{itemize}
  \item We propose a pipeline, NEAS, searching for diverse models under certain resource constraints. Our approach could search for both homogeneous and heterogeneous ensemble models.
  \item We design a new metric, diversity score, to guide the shrinking process of search space. We evaluate its superiority on supernet training and the performance of searched models by enormous experiments.
  \item We propose a layer-sharing strategy to reduce the complexity of ensemble models and enlarge the search space to search for an optimal split point. 
  \item We compare the searched architectures to state-of-the-art NAS methods on the image classification task and achieve state-of-the-art results. Furthermore, we evaluate our searched model on the downstream object detection task, showing their generalization ability.

\end{itemize}
\section{Related works}
\noindent\textbf{Neural Architecture Search.} NAS has shown its superiority to manual-crafted networks on varied vision tasks, such as image classification \cite{tan2020efficientnet}, semantic segmentation \cite{liu2019auto} and object detection \cite{chen2019detnas}. Early NAS approaches search the architectures using either reinforcement learning \cite{zoph2016neural, zoph2018learning, zhong2018practical} or evolution algorithms \cite{real2019regularized, suganuma2017genetic}. However, these approaches require training thousands of architecture candidates from scratch, leading to unaffordable computation overhead. Most recent works resort to the weight sharing strategy to amortize the searching cost. Those approaches train a single over-parameterized supernet and then share the weights across subnets. They could be further categorized as two types: path-based \cite{guo2020single, chu2019fairnas, chu2020mixpath} and gradient-based methods \cite{liu2018darts, chen2019progressive, xu2020pcdarts}. Path-based methods sample paths in each iteration to optimize the weights of supernets. Once the training process is finished, the subnets can be ranked by the shared weights. On the other hand, gradient-based methods relax the discrete search space to be continuous, and optimize the search process by the efficient gradient descent.

Recent work NES \cite{zaidi20} proposes to search diverse architectures to form robust ensembles against distributional shift of the data. NES shows that the ensemble of different architectures performs better than fixed architectures and provides a new regularized evolution search algorithm for ensemble architecture search. Since NEAS searches for efficient and effective models and NES searches for models without considering the computation cost, we do not compare NES in our work.

\noindent\textbf{Ensemble Learning.} Ensemble methods are widely used to build stronger and robuster models than single models. \cite{zhou2012ensemble, dropout,dropconnect, huang2017snapshot, zhou2002ensembling, simonyan2014very}. Strategies for building ensembles could be mainly divided into two categories. The first ones train different models independently and then apply ensemble methods to form a more robust model, such as boosting, bagging, and stacking \cite{zhou2002ensembling}. The other methods train only one model with specific strategies to achieve implicit ensemble \cite{zhou2012ensemble, dropout,dropconnect, huang2017snapshot}. Different from the above methods, we perform explicit ensemble without separate training and search for diverse model architectures to build ensemble models with great feature expression ability.

\noindent\textbf{Search Space Shrinking.} Recent works have shown search space shrinking is effective in boosting the ranking ability of NAS methods, especially when the search space is huge. \cite{hu2020anglebased, li2019improving, noy2019asap, nayman2019xnas}. These methods could be classified into different types according to their evaluation metrics. There are three basic types: accuracy-based, magnitude-based, and angle-based metrics. For example, PCNAS \cite{li2019improving} drop unpromising operators layer by layer using accuracy and shows that it improves candidate networks' quality. AngleNAS \cite{hu2020anglebased} uses the angles between weights of models to guide the search process. However, existing shrinking techniques only consider operators independently. Therefore, they can't directly adapt to search for ensemble models. We design a new metric considering both the performance of single operators and the diversity across them.
\vspace{-0.5mm}
\section{Approach}
\begin{figure*} 
\centering
\includegraphics[width=1\textwidth]{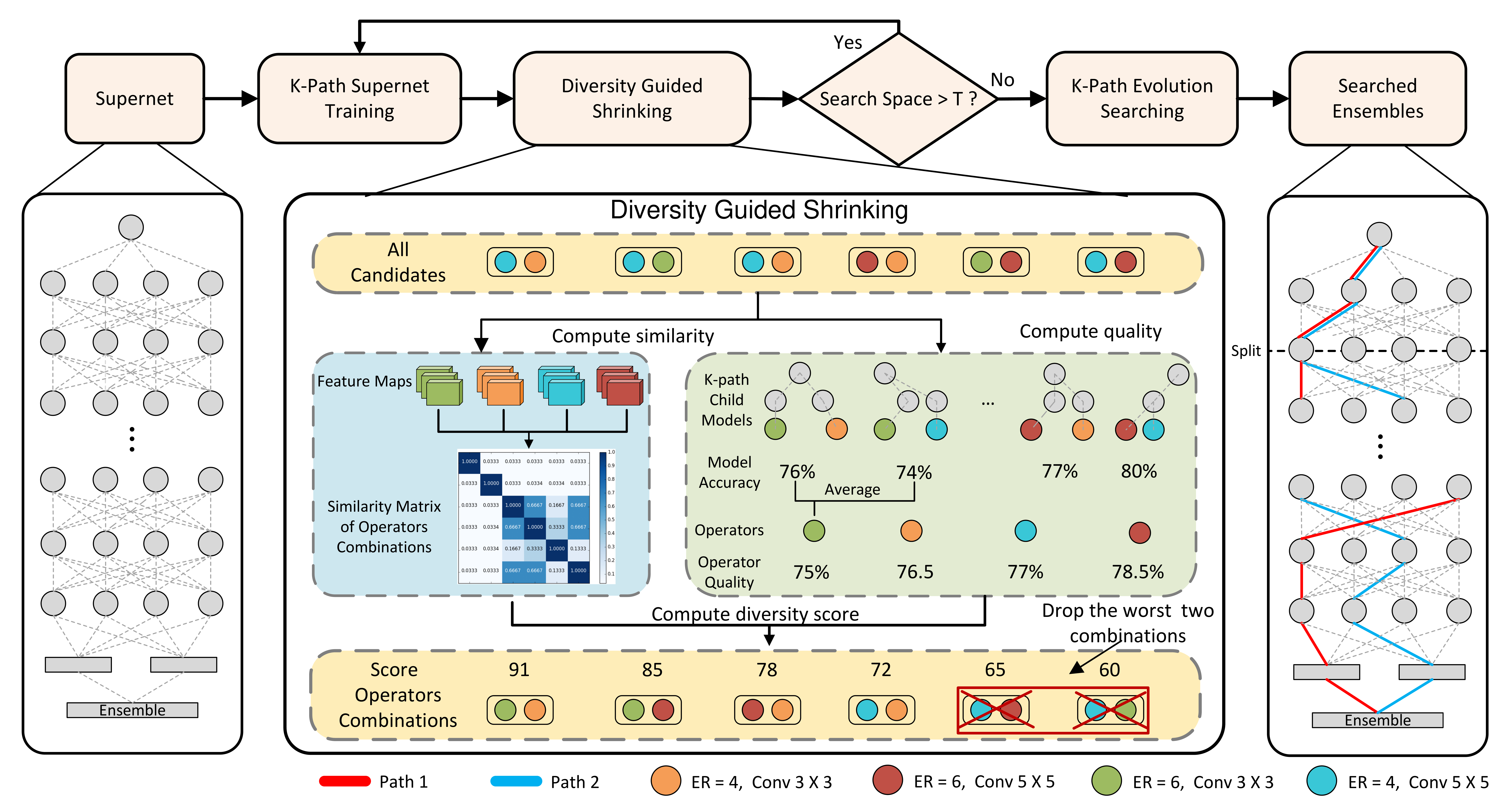} 
\caption{NEAS contains mainly two steps: K-path Supernet Training with Diversity-Guided Shrinking, and K-path Evolution Searching. It takes the search space as the input and outputs an ensemble model with shared shallow layers. We set the number of paths in the searched models to 2 and choice operators to 4 for explanation. The overlapping upper lines in the right graph indicate that the two paths share the first two layers. They then branch to two different paths. ER means the expansion ratio for the mobile inverted residual block.}
\label{Fig.overview} 
\end{figure*}

In Section \ref{3.1}, we give the formulation of NEAS. In Section \ref{3.2}, we present the definition of the diversity score and the space shrinking pipeline. In Section \ref{3.3}, we introduce the layer sharing mechanism and the new search dimension \textit{Split Point}. In Section \ref{3.4}, we give the detailed pipeline of NEAS which allows to search under different resource constrains.
The overall framework is visualized in Fig.~\ref{Fig.overview}.

\subsection{NEAS Formulation} \label{3.1} 
Given the search space $\Omega$ of single deep neural networks, denote $\mathcal{A}=\{\phi_k\in\Omega: k=1,\dots,K\}$ as a set of $K$ architectures with corresponding parameters $\mathcal{W}=\{\omega_k:k=1,\dots,K\}$, $\Phi(\cdot; \mathcal{A},\mathcal{W})$ as the ensemble model, and $\mathcal{S}= \Omega^K$ as the search space of ensemble models. The goal of NEAS is to find an optimal architectures set $\mathcal{A}^*$ that maximizes the overall validation accuracy. To reduce the search cost, we constrain $\Omega$ to a certain architecture family, specifically, the subnetworks induced by a predefined supernet. 
In our work, we specify $\Phi(\cdot; \mathcal{A},\mathcal{W})$ as:
\begin{equation}
    \Phi(\cdot;\mathcal{A}, \mathcal{W}) = \sum\limits_{i = 1}^{K}\frac{\phi_i(\cdot;w_i)}{K}.
\end{equation}

We then formulate NEAS as a two-stage optimization problem like other one-shot methods (\eg,~\cite{guo2020single}). The first-stage is to optimize the weight of the supernet by: 
\begin{equation}
    W_{\mathcal{S}} = \argmin_{W} \mathcal{L}_{\rm train} (\Phi(\cdot;\mathcal{A}, W(\mathcal{A}))), 
\end{equation}
where $\mathcal{L}_{\rm train}$ is the loss function on the training set, $W(\mathcal{A})$ means architectures in ${\mathcal{A}}$ inherit weights from $W$.

This step is done by uniformly sampling an ensemble architecture $\Phi$ from $\mathcal{S}$ and performing backpropagation to update the weight of the corresponding blocks in the supernet for each iteration. Please refer to Section \ref{3.4} for details. 

The second step is to search for an optimal architecture set $\mathcal{A}^*$ via ranking the performance based on learned weight $W_{\mathcal{S}}$ of supernet, which is formulated as
\begin{equation}
\begin{aligned}
    & \mathcal{A}^* = \argmax_{\mathcal{A}} {\rm ACC}_{\rm val} (\Phi(\cdot;{\mathcal{A}}, W_{\mathcal{S}}({\mathcal{A}}))),\\
    & {\rm s.t.} \quad \sum_i^{K}{g_i(\phi_i)} < C,
\label{eq:step2}
\end{aligned}
\end{equation}
where $g$ and $C$ are the resource computation functions and the resource constraints. Typical constraints include FLOPs, parameters size, and run-time latency. 

Since it is difficult to enumerate all ensemble architectures for evaluation, we resort to a specific $K$-path evolution algorithms to find the most promising one. The details are presented in \textit{Appendix B} and Section \ref{3.4}.

\subsection{Diversity-Guided Search Space Shrinking} \label{3.2}
Since we search directly for the ensemble models, the search space for each layer increases exponentially from $N$ to $A_N^K=\frac{N!}{(N-K)!}$ compared with single path methods, where $N$ is number of the alternative operators for each layer. The large search space causes inefficiency search and supernet optimization problem. Search space shrinking is a feasible solution to alleviate the problem by discarding inferior operators progressively with a specific metric. Since diversity plays a key role in building a robust ensemble model, we design a new metric to explicitly quantify the diversity across operators inspired by fixed-size determinantal point processing ($K$-DPP) \cite{kulesza2011k}, a  popular sampling model with great ability to measure the global diversity and quality within a set. In the following section, we first define the diversity score of an operator combination and then present the diversity-guided search space shrinking pipeline.

\noindent\textbf{Definition of Diversity Score.}\quad Assume we have an ensemble model $\Phi(.;\mathcal{A}, W(\mathcal{A}))$, wherer $\mathcal{A} = \{\phi_1, \phi_2, \cdots, \phi_K\}$ consisting of $K$ different paths. Since we fix the depth of the search space, we can slice $\mathcal{A}$ into operator combinations by layer. Then, $\mathcal{A} $ can be reshaped as $ \{h_m| h_m = (o_{1,m}, o_{2,m}, \cdots, o_{K,m}), m=1,2,\cdots, d\}$, where $m$ and $d$ are the index of the layer and number of total layers, $o_{i,m}$ denotes the operator on layer $m$ of the path $i$. Now our goal changes to find the optimal operator combination for each layer.

Given that layer $m$  has $N$ alternative operators $O = O_{1,m}, O_{2,m},\cdots, O_{N,m}$, we construct a DPP kernel $L \in \mathbb{R}^{N \times N}$ for layer $m$ as:
 \begin{equation}
L_m = {\rm diag}(r^m) \cdot S^m \cdot {\rm diag}(r^m),
\end{equation}
where the kernel is formed by two components: a similarity matrix $S^m \in \mathbb{R}^{N \times N}$ and a quality matrix $r^m \in \mathbb{R}^N$.

Let $v_{1}, \cdots, v_{K}$ denote the feature maps output from the $K$ different paths $\phi_{1}, \cdots , \phi_{K}$ of the ensemble model $\Phi(.;\mathcal{A}, W(\mathcal{A}))$. We define the similarity $S_{i,j}^m$ of two operators $O_{i,m}$ and $O_{j,m}$ as expected the similarity between paths that contains the two operators, respectively:
\begin{equation}
S_{i,j}^m = \mathbb{E}_{\mathcal{A}\subseteq \mathcal{S}}\Big(\sum_{p,q}{\mathbb I(i,j,p,q) \exp (-\beta \|v_{p} - v_{q}\|_2)}\Big),
\label{eq:similarity}
\end{equation}
where $1\leq p,q\leq K$, $\beta$ is a scaling factor, and the indicator function is defined as:
\begin{equation}
\mathbb{I}(i,j,p,q)=\left\{
\begin{aligned}
1, &  & O_{i,m} \in \phi_{p}, \\
0, &  & O_{j,m} \in \phi_{q}.\\
\end{aligned}
\label{eq:indicator}
\right.
\end{equation}

The quality of operator $O_{i,m}$ is computed by taking expected accuracy of paths containing it. The formal definition is:
\begin{equation}
r_{i}^m = \gamma ~ \mathbb{E}_{\mathcal{A}\subseteq \mathcal{S}} \left({\frac{\sum_{\phi_{q}|O_{i,m}\in \phi_{q}}{\rm{ACC}_{train'}(\phi_{q})}}{\rm{\#}\{\phi_{p}|O_{i,m}\in \phi_{p}\}}} \right),
\label{eq:quality}
\end{equation}
where $\phi_{q},\phi_{p}\in \mathcal{A}$, $\rm{ACC}_{train'}$ is the accuracy evaluated on a small part of training dataset. 

In practice, we do not calculate the exact expectation of similarity matrix and quality matrix. Instead, we randomly sample a finite number of ensemble models and use the mean as an approximate of the expectation.

The diversity score of a certain operator combination $h_m$ of layer $m$ is defined as following:
\begin{equation}
Score(h_m)  = {\rm det}(L_m^y),
\label{eq:score}
\end{equation}
where $L_m^y$ is the submatrix of $L_m$ that contains all operators of $h_m$. The trade-off between similarities and accuracy is controlled by the hyperparameter $\gamma$.

According to the the definition of diversity score, we have the following property: \\
For $h_m$ and $h_m'$ that are different by only the $i_{th}$ operator, if \ $S_{i,j}^m < S_{i',j}^m$ for $j = 1,2,\cdots,K$ and $r_i^m > r_{i'}^m$, then
\begin{equation}
\begin{aligned}
&Score(h_m) > Score(h_m^{'}).
\end{aligned}
\end{equation}
This property suggests that the metric will drop similar and unpromising operator combinations while keep diverse and accurate operator combinations. We refer to \textit{Appendix A} for a proof.

\begin{algorithm}[!t]
	\renewcommand{\algorithmicrequire}{\textbf{Input:}}
	\renewcommand{\algorithmicensure}{\textbf{Output:}}
	\caption{Diversity-Guided Search Space Shrinking}
	\label{alg:DA}
	\begin{algorithmic}[1]
		\REQUIRE 
	    A search space $\mathcal{S}$, threshold of search space size $\mathcal{T}$, number of operators dropped out each shrinking $k$, supernet $\mathcal{G}$, number of ensembles sampled each shrink $Z$, training epochs between each shrink $E$.
		\ENSURE A shrunk search space $\tilde{\mathcal{S}}$.
		\STATE Let $\tilde{\mathcal{S}}=\mathcal{S}$\;
		\WHILE {$|\tilde{\mathcal{S}}| >\mathcal{T} $}
		\STATE Training the supernet $\mathcal{G}$ for $E$ epochs following Section \ref{3.4};
        \STATE Sample $Z$ ensemble models $\Phi_1, \Phi_2, \cdots, \Phi_Z$ randomly;
		\STATE Compute diversity score of each operator combination from $\tilde{\mathcal{S}}$ using Eq. \ref{eq:similarity},\ref{eq:quality},\ref{eq:score};
		\STATE Removing $k$ operator combination from $\tilde{\mathcal{S}}$ with the lowest $k$ scores\;
		\ENDWHILE
	\end{algorithmic} 
\end{algorithm}

\vspace{.5mm}\noindent\textbf{Diversity-Guided Search Space Shrinking.} Based on the diversity score, we present Algorithm \ref{alg:DA} to describe the diversity-guided search space shrinking pipeline shown in middle of Fig.~\ref{Fig.overview}. Note that during the shrinking process, at least one operator combination is preserved, since our method does not change the connectivity of the supernet.

\subsection {Layer Sharing Among Ensemble Components} \label{3.3}
\begin{figure}
\centering
\includegraphics[width=0.45\textwidth]{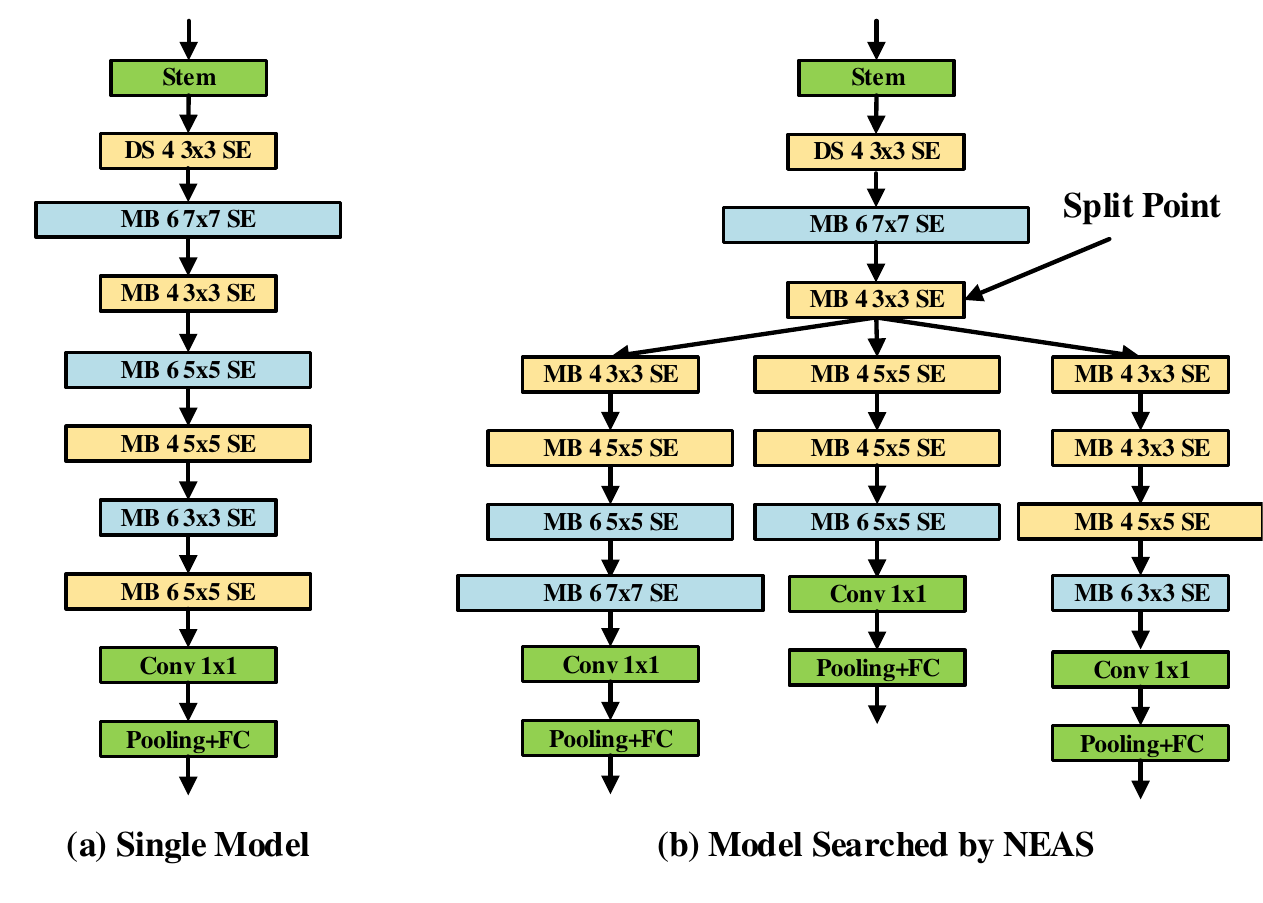} 
\caption{(a) The architecture searched by classical NAS methods (\eg, \cite{guo2020single, chu2019fairnas}). (b) The architecture searched by NEAS. Different color means different expansion ratio while the length of the block represents the kernel size.}
\label{Fig.structure} 
\vspace{-3mm}
\end{figure}

The challenge of potential massive complexity of searched ensemble models is handled by the layer sharing mechanism. This mechanism is inspired by several recent studies \cite{kornblith2019similarity, morcos2018insights, raghu2017svcca}. These works find that both the same neural architectures with different initialization and different architectures learn similar features in their lower layers. Therefore, we consider to share the shallow layers of different ensemble components. We propose to search for diverse ensemble components with shared shallow layers and different deep layers to reduce the computation cost. To automatically find which layers should be shared, we design a new search dimension called \textit{split point}. The split point defines where the ensemble model will have heterogeneous architectures. It also handles the trade-off between diversity and computation constrain. A comparison between the architectures searched by NEAS and other NAS methods such as \cite{guo2020single, howard2019searching} is presented in Fig.~\ref{Fig.structure}.

\subsection{Neural Ensemble Architecture Search} \label{3.4}
As state in Section \ref{3.1} and in Fig.~\ref{Fig.overview}, NEAS includes two sequential phases: K-path supernet training with diversity-guide search space shrinking, and K-path evolution search.

\vspace{.5mm}\noindent\textbf{Phase 1: K-Path Supernet Training with Diversity-Guide Search Space Shrinking.}\quad
For each training iteration, 
an ensemble model $\Phi(.;\mathcal{A}, W(\mathcal{A}))$ is randomly sampled.
In specific, we randomly sample the split point $s$, the architecture of sharing layers $\mathcal{A}_{sharing} = \{o_1, o_2,\cdots, o_s\}$, and the operator combinations $\mathcal{A}_{split} = \{h_{s+1} , h_{s+2}, \cdots, h_{d}\}$ for the rest of layers from the shrunk search space. The loss $\mathcal{L}_i$ of each path $\phi_{i}$ is computed independently while the backpropagation is performed using the combined loss $\mathcal{L} = \sum^{K}_{i}\mathcal{L}_i$ to update the weights of corresponding blocks in the supernet. Following this updating process, the whole network is still trained in an end-to-end style. After training the supernet for several epochs, we follow the steps in Algorithm~\ref{alg:DA} to shrink the search space. The shrinking and training are conducted alternatively.

During inference, these selected paths make predictions independently, and our ensemble network's output is the average of predictions from all paths.

\vspace{.5mm}\noindent\textbf{Phase 2: K-Path Evolution Search.} After obtaining the trained supernet, we perform evolution search on it to obtain an optimal ensemble model. These models are evaluated and picked according to the manager of the evolution algorithm. It is worth noting that, before evaluating an ensemble model, we first need to recalculate the batch normalization (BN) statistics for each block. This is because, during the supernet training, the BN statistics of different blocks are optimized simultaneously. These statistics are usually not applicable to the subnets. We randomly extract a part of the ImageNet training set to recalculate the BN statistics. 
 
At the beginning of the evolution search, we pick $N_{\rm seed}$ random architecture as seeds. The top $k$ architectures are picked as parents to generate the next generation by crossover and mutation. In one crossover, two randomly selected candidates are picked and crossed to produce a new one during each generation. We drop the architecture got by crossover if the corresponding architecture is not in the shrunk search space or exceeds the FLOPs constraint. In one mutation, a candidate mutates its split point with a probability $P_s$. If the split point increases, the number of sharing layers increases with the same number. We randomly pick one path and move its corresponding architectures to the sharing architecture. Otherwise, if the split point decreases, we cut the sharing architecture and add it to each path's architecture. At last, the candidate mutates its layers with a probability of $P_m$ to produce a new candidate. It is worth noting that the operation combinations are only picked from the shrunk search space. We perform crossover and mutation several times to generate new
candidates. We generate some random architectures after crossover and mutation to meet the given population demanding. We provide the detailed algorithm in the \textit{Appendix B}.

\section{Experiment}
In this section, we first give details of our search space and implementation. We then present ablation studies dissecting our method, followed by a comparison with previous state-of-the-art NAS methods. At last, we evaluate the generalization ability and robustness of the searched architecture on COCO object detection benchmark. \begin{table*}[t!]
	\vspace{-0.3cm}
	\caption{Comparison of state-of-the-art NAS methods on ImageNet. $\dagger$: TPU days, $\star$: reported by \cite{guo2020single}, $\ddagger$: searched on CIFAR-10, "-" means not reported. $\diamondsuit$: Tested on \textit{NVIDIA GTX 1080Ti}.} 
	\centering
	\small
    {
		\begin{tabular}{c|c||ccc||cccc} 
			\toprule[1.5pt]
			\multicolumn{1}{c|}{}& {\multirow{2}{*}{Methods}} & Top-1 & Top-5 & FLOPs  & {\multirow{2}{*}{Memory cost}} & Superne train & Search cost & {\multirow{2}{*}{Retrain epochs}}\\
			\multicolumn{1}{c|}{}& {} & (\%) & (\%) & (M)  & {}  & (GPU days) & (GPU days)  &{}\\
			
			\midrule[1.5pt]
			\multirow{8}{*}{\rotatebox{90}{200 -- 350M}}   & $\text{MobileNetV3}_{Large1.0}$ \cite{howard2019searching}& 75.2 & - & 219&single path & $288^\dagger$ &- & 150 \\
			& OFA \cite{cai2019once} &76.9 & - &230 &two paths&53&2& - \\
			& $\text{MobileNetV2}$ \cite{sandler2018mobilenetv2} &  72.0& 91.0& 300 & - & - & - & - \\
			& MnasNet-A1 \cite{tan2019mnasnet}& 75.2 & 92.5& 312 &single path & $288^\dagger$ &- & 350 \\
			& FairNAS-C \cite{chu2019fairnas}&  74.7&92.1&321 &single path&10  &2 & - \\
			& FBNetV2-L1 \cite{wan2020fbnetv2} & 77.2 & - & 325 & - & - & - & 400 \\
			& SPOS \cite{guo2020single}& 74.7&-&328&single path&$\text{12}^{\diamondsuit}$&$<1$ & 240 \\

			& \textbf{NEAS-S (Ours)} &\textbf{77.9} & \textbf{93.9} &314 &$K$ paths&12  &$<1$ & 350 \\
			
			\midrule[1.5pt]	
			\multirow{6}{*}{\rotatebox{90}{350 -- 500M}} & GreedyNAS-A \cite{you2020greedynas}&77.1 &93.3&366 &single path&7&$<1$& 300 \\
			& EfficientNet-B0 \cite{tan2020efficientnet} & 77.1 & 93.3& 390 & - & -& -& 350  \\
			& FBNetV2-L2 \cite{wan2020fbnetv2}  & 78.2 & - & 422 & - & - & - & 400 \\
			&ProxylessNAS\cite{cai2018proxylessnas}&75.1& - &465&two paths&$15^\star $&-& 300  \\
			& Cream-M \cite{Cream}  & 79.2 & 94.2 & 481 & two paths & 12 & 0.02 & 500 \\
			& \textbf{NEAS-M (Ours)} &\textbf{79.5} & \textbf{94.6} &472 &$K$ paths&12 &$<1$& 350 \\

			\midrule[1.5pt] 
			& DARTS \cite{liu2018darts} & 73.3& 91.3& 574&whole supernet &$4^\ddagger$&-& 250 \\  
			\multirow{4}{*}{\rotatebox{90}{500 -- 600M}}  & BigNASModel-L \cite{yu2020bignas} &79.5 & - &586&two paths&$96^\dagger$& - &-\\
			& $\text{OFA}_{Large}$ \cite{cai2019once} &80.0 & - &595&two paths&53&2& - \\
			& DNA-d \cite{li2020block} &78.4 & 94.0 &611&single path&24&0.6& 500 \\
			& EfficientNet-B1 \cite{tan2020efficientnet} & 79.2 & 94.5& 734 & - & -  & - & 350\\ 
			& \textbf{NEAS-L (Ours)} &\textbf{80.0} & \textbf{94.8} &574& $K$ paths&12&$<1$& 350 \\
			\bottomrule[1.5pt]
		\end{tabular} 
	}
	\vspace{-0.2cm}
	\label{tab:sota}
\end{table*}

\subsection{Implementation Details}

\noindent\textbf{Search Space.} Consistent with previous NAS methods \cite{guo2020single,chu2019fairnas,tan2020efficientnet}, our search space includes a stack of mobile inverted bottleneck residual blocks (MBConv). We also add squeeze-excitation modules to each block following EfficientNet \cite{tan2020efficientnet} and MobileNetV3 \cite{howard2019searching}. For details, there are 7 basic operators for each layer, including MBConv with kernel sizes of {3,5,7}, expansion rates of {4,6} and skip connect for elastic depth. The split point space is set to range (9, 20) to handle different complexity constrains. In total we have $7^{20K} \times 12 \geq 7 \times 10^{33}$ ($K\geq2$) architectures, which is much larger than most NAS methods. A more detailed description of search space could be found in \textit{Appendix A}.

\noindent\textbf{Supernet Training.} We train the supernet for 120 epochs using the settings similar to SPOS \cite{guo2020single}: SGD optimizer with momentum 0.9 and weight decay 4e-5, initial learning rate 0.5 with a linear annealing. The shrinking process is conducted every 20 epochs. The number of operators dropped each time is empirically set to 20. $\beta$ in the computing the similarity matrix is set to 1e-3 according to experimental results.

\noindent\textbf{Evolution Search.} We set the population $N_{\rm seed}$ of evolution search to 50 with the size of top candidates pool $k$ equals to 10. The number of generations is 20. $P_s$ and $P_m$ are both 0.1. The number of candidates performs mutation and crossover are set to 25 in each generation. We recalculate the BN statistics on a subset of ImageNet.   

\noindent\textbf{Retrain.} We retrain the discovered architectures for 350 epochs on ImageNet using similar settings as EfficientNet~\cite{tan2020efficientnet}: RMSProp optimizer with momentum 0.9 and decay 0.9, weight decay 1$\mathrm{e}${-5}, dropout ratio 0.2, initial learning rate 0.064 with a warmup in the first 10 epochs and a cosine annealing. AutoAugment \cite{cubuk2019autoaugment} and exponential moving average are also used for training. We retrain the models with a batch size of 2,048 on 16 Nvidia Tesla V100 GPUs.
\begin{table}
\caption{Comparison of different shrink metrics. Baseline means no search space shrinking during the supernet training. $\dag$: average accuracy use weight inherits from supernet. The accuracies are evaluated on ImageNet.}
\small
\centering
\vspace{0.5mm}
\resizebox{\columnwidth}{!}{
\begin{tabular}{c|cccc} 
\toprule[1pt]
Metric & Kendall Tall & Top-1 (\%) &Top-5 (\%) & $\text{Top-1}^\dag$ (\%)\\
\midrule[1pt]
Baseline & 0.45 & 77.3 & 93.3 & 67.8 \\ 
Accuracy & 0.42 & 77.2 & 93.2 & 67.2 \\ 
Diversity & 0.65 & 77.9 & 93.9 & 68.3\\ 
\bottomrule[1pt]
\end{tabular}
}
\label{tab.kendall}
\end{table}
\subsection{Ablation Study}
 
\noindent\textbf{Effectiveness of Diversity Score.} We set the baseline as NEAS without diversity-guided shrinking. In addition, we compare the diversity score with the accuracy metric to further verify its efficacy. Since the accuracy-based methods only consider the accuracy of single operators in each layer. We adapt the definition of accuracy to the accuracy of operator combinations. Other methods like Angle-based metric can not easily adapt to search for ensembles.

We first perform correlation analysis to evaluate whether the training process with diversity shrinking can improve the ranking ability of supernet. We randomly sample 30 subnets and calculate the rank correlation between the weight sharing performance and the true performance of training from scratch. Training many such subnets on ImageNet is very computationally expensive. We follow the setting of Cream \cite{Cream}, which constructs a subImageNet dataset consisting of 100 classes randomly sampled from ImageNet. Each class has 250 training images and 50 validation images. We use Kendall Tau to show the ranking capacity of supernet. The second column of Table~\ref{tab.kendall} suggests that our diversity score effectively helps supernet to rank the ensemble architectures in the supernet.

We also retrain the searched architectures by the three methods under the same FLOPs constraint. The top-1 and top-5 accuracy results on the ImageNet dataset are shown in the third and fourth columns in Table~\ref{tab.kendall}. We could see that the diversity-guided shrinking is 0.6\% better than the baseline and 0.7\% better than the accuracy-based method. We further compare the average accuracies of the architectures in the last generation of evolution search, displaying in the fourth columns. Our diversity-guided shrinking surpass the baseline and accuracy-based method by 0.5\% and 1.1\% top-1 accuracy on ImageNet in the supernet. The results suggest that the diversity score helps remove unpromising candidates and enhance the convergence of supernet.

\begin{table}[!t]
\caption{Comparison of architectures with homogeneous (homo) and heterogeneous (hetero) paths. Both architectures have two paths. The numbers in the columns 2,3,4 are the top-1 accuracies.}
\vspace{0.5mm}
\resizebox{\columnwidth}{!}{
\begin{tabular}{c|cccc} 
\toprule[1pt]

 & Path 1 (\%) & Path 2 (\%) & Ens.(\%) & FLOPs(M)\\
\midrule[1pt]
homo (first)  & 79.0 & 79.1 & 79.4 & 566\\
homo (second)  & 79.2 & 79.3 & 79.6 & 586\\
hetero (baseline) & 78.9 & 79.0 & 80.0 & 574\\ 
\bottomrule[1pt]
\end{tabular}
\label{tab.why_ensemble}
}
\end{table}

\begin{table}[!t]
\caption{Comparison of split point in different searched models. Baseline: ensemble model (2 models) with no shared layers. Top-1 and Top-5 represents the top-1 and top-5 accuracy on ImageNet.}
\vspace{0.5mm}
\resizebox{\columnwidth}{!}{
\begin{tabular}{c|cccc} 
\toprule[1pt]
Model & Split point & Top-1 (\%) & Top-5 (\%) & FLOPs (M)\\
\midrule[1pt]
NEAS-L & 16 & 80.0 & 94.8 & 574\\ 
NEAS-M & 16 & 79.5 & 94.6 & 472\\ 
NEAS-S & 20 & 77.9 & 93.9 & 314\\ 
baseline & - & 78.5 & 94.2 & 605\\
\bottomrule[1pt]
\end{tabular}
\label{tab:weight_sharing}
}
\vspace{-2mm}
\end{table}

\begin{figure}[!t]
\centering
\includegraphics[width=0.4\textwidth]{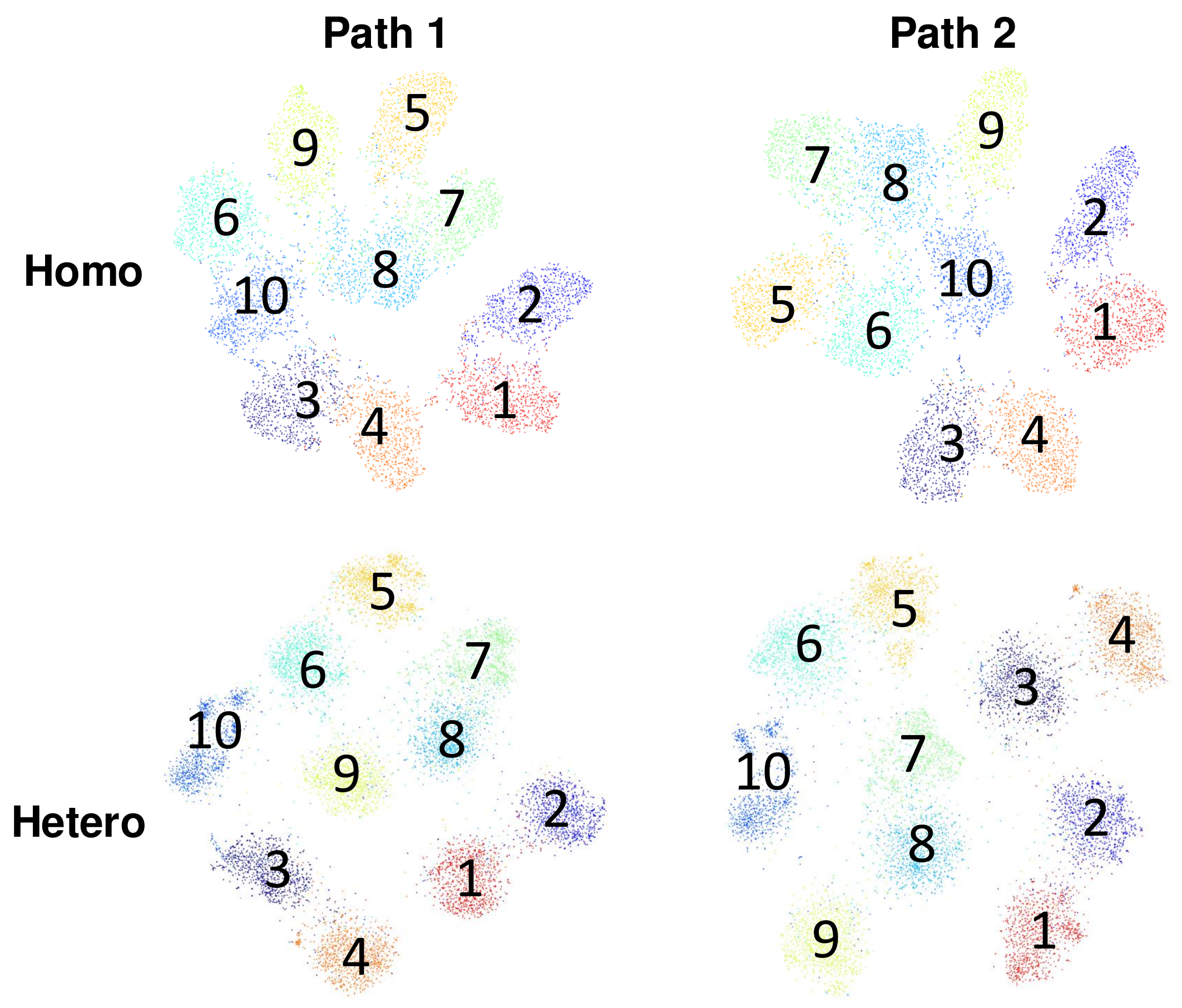} 
\caption{t-SNE visualization on the final hidden features
of two different ensemble model. 
The first row denotes the model with two homogeneous path, while the second has two heterogeneous paths. The inputs are the test set of CIFAR-10.}
\label{Fig.features} 
\vspace{-3mm}
\end{figure}

\noindent\textbf{Impact of Heterogeneous Path Architectures.}
Ensembling models of homogeneous (homo) architectures are known to be an effective way of building powerful models \cite{lakshminarayanan2017simple}. We here compare the ensemble models of homogeneous and heterogeneous architectures to show the importance of heterogeneous ensemble. We use our searched two-path ensemble model as the baseline. Then we mirror one path of the searched architecture to form two homogeneous ensemble models for comparison. Fig.~\ref{Fig.features} gives the visualization of the final hidden features of baseline and the homogeneous network fine-tuned on CIFAR-10. We can see that the homo paths have a similar feature distribution. However, the hetero paths have varied feature distribution and a clearer margin between the clusters.

In table \ref{tab.why_ensemble}, we compare the performance of these three models on ImageNet. This table shows an interesting fact that even the stand-alone performance of homo paths are better than hetero. However, the performance of the homo ensemble is worse than the heterogeneous one, indicating that the two paths of the searched model are complementary.

\begin{table}[!t]
\vspace{3mm}
\caption{Impact of number of paths predefined for NEAS. Top-1 and Top-5 represents the top-1 and top-5 accuracy on ImageNet.}
\centering
\vspace{0.5mm}
\begin{tabular}{ ccccc } 
\toprule[0.75pt]
\#paths & Top-1 (\%) & Top-5 (\%) & FLOPs (M) \\
\midrule[0.75pt]  
2 & 80.0 & 94.8 & 574  \\ 
3 & 79.5 & 94.6 & 564  \\ 
5 & 78.5 & 94.1 & 570\\ 
\bottomrule[0.75pt]
	\label{tab:paths}
\end{tabular}
\vspace{-5mm}
\end{table}

\begin{table*}[tp]
    \small
	\caption{\small{Object detection results of various drop-in backbones on COCO \texttt{val2017}. Top-1 accuracies are on ImageNet. $\dag$: reported by~\cite{chu2019fairnas}}.}
	\small
	\vspace{-0.2cm}
	\begin{center}
		{
		\begin{tabular}{*{1}{l}*{8}{c}}
				\toprule[0.75pt]
				Backbones & FLOPs (M)  & AP (\%)& AP$_{50}$ & AP$_{75}$ & AP$_S$ & AP$_M$ & AP$_L$ &Top-1 (\%) \\
				\midrule[0.75pt]
				$\text{MobileNetV3}^\dag$ \cite{howard2019searching} & 219 & 29.9 & 49.3 & 30.8 & 14.9 & 33.3 & 41.1 & 75.2\\
				$\text{MobileNetV2}^\dag$ \cite{sandler2018mobilenetv2} & 300& 28.3 & 46.7 & 29.3 & 14.8 & 30.7 & 38.1 & 72.0\\
				FairNAS-C \cite{chu2019fairnas} &325& 31.2 & 50.8 & 32.7 & 16.3 & 34.4 & 42.3 & 76.7\\
				$\text{MnasNet-A2}^\dag$ \cite{tan2019mnasnet} & 340 & 30.5 & 50.2 & 32.0 & 16.6 & 34.1 & 41.1 & 75.6\\
				$\text{MixNet-M}^\dag$ \cite{tan2019mixconv} & 360 & 31.3& 51.7 & 32.4& 17.0 & 35.0 & 41.9 & 77.0   \\
				$\text{SPOS}^\dag$ \cite{guo2020single} & 365  & 30.7 & 49.8 & 32.2 & 15.4 &33.9 & 41.6 & 75.0\\
				\midrule[0.75pt]
				\textbf{NEAS-S} &314 & \textbf{33.0} & \textbf{53.3} & \textbf{34.4} & \textbf{17.9} & \textbf{36.2} & \textbf{43.8} & \textbf{78.0} \\
				\bottomrule[0.75pt]
			\end{tabular}
		}
	\label{tab:det_retina}
	\end{center}
\vspace{-0.4cm}
\end{table*}
\noindent\textbf{Impact of Layer Sharing.} Layer sharing plays a significant role in reducing the complexity of an ensemble model. Here, we explore the effectiveness of layer sharing. The baseline is the ensemble model with no shared layers searched by our method. In Table~\ref{tab:weight_sharing}, we could see that layer sharing will help to reduce the complexity of ensemble models largely while keeping outstanding performance. Besides, we observed that in our searched models, the larger model attempts to share fewer layers. One reason could be that the feature expression ability of stand-alone paths in larger models is already strong since it is more complicated. Therefore, they prefer to share fewer layers and get more diverse paths.

\noindent\textbf{Impact of Number of Paths for Ensemble.}
The number of paths $K$ used to form the ensemble model is a hyperparameter we define at first. We compare the performance of the searched model under the mobile setting ($\leqslant$600M FLOPs) using different $K$. From Table~\ref{tab:paths}, we can see that when the number of paths is equal to 2, we achieve the best results. One likely reason could be that if a network has too many paths, each path's stand-alone feature expression ability decreases a lot due to complexity constraints.

\begin{figure} [!t]
\centering
\includegraphics[width=0.45\textwidth]{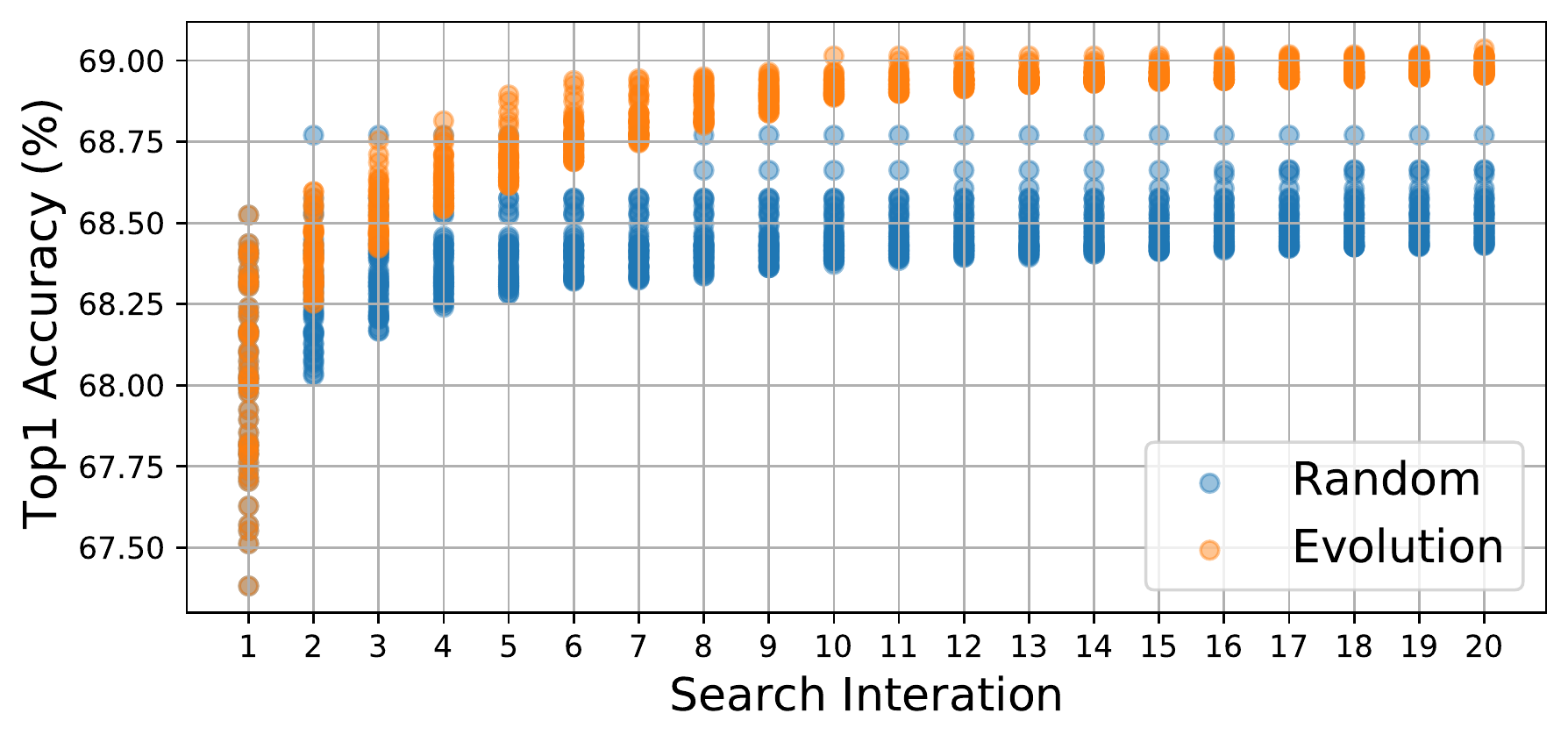} 
\caption{Random search versus evolution algorithm.}
\label{Fig.evo}
\vspace{-3mm}
\end{figure}

\noindent\textbf{Impact of Search Algorithm.}
Random search is known to be a competitive baseline in NAS methods. We compare random search with evolution search to evaluate the effectiveness of evolution search. We demonstrate the performance of architectures using the weights inherited from supernet on the validation dataset during the search. Top 50 candidates until the current iteration are depicted at each iteration. Fig.~\ref{Fig.evo} illustrates that evolution search is better for searching on supernet.  

\subsection{Comparisons with State-of-the-Art Methods}

Table~\ref{tab:sota} presents the comparison of our method with state-of-the-arts under mobile settings on ImageNet. It shows that when considering models with FLOPs smaller than 600M, our method consistently outperforms the recent MobileNetV3 \cite{howard2019searching} and EfficientNet-B0/B1 \cite{tan2020efficientnet}. In particular, NEAS-L achieves 80.0\% top-1 accuracy with only 574M FLOPs, which is 160M FLOPs smaller and 0.8\% better than EfficientNet-B1. NEAS-M obtains 79.5\% top-1 accuracy with 472M FLOPs. NEAS-S achieves 77.9\% accuracy using only 314M FLOPs, which is 0.8\% better and 19\% smaller than EfficientNet-B0. We also provide results of other state-of-the-art NAS methods in Table~\ref{tab:sota}. It is worth noting that some NAS methods like OFA \cite{cai2019once}, BigNAS \cite{yu2020bignas}, DNA \cite{li2020block} use knowledge distillation to boost the training process and also improve the accuracy of searched models. However, even compared with these methods, our searched ensemble architectures, which do not use knowledge distillation, still achieve superior performance.

\subsection{Generalization Ability and Robustness}
To further evaluate the generalization ability of the architectures found by NEAS, we transfer the architectures to the downstream COCO \cite{lin2015microsoft} object detection task. We use the NEAS-S (pre-trained 500 epochs on ImageNet) as a drop-in replacement for the backbone feature extractor in RetinaNet \cite{lin2017focal} and compare it with other backbone networks. We perform training on the train2017 set (around 118k images) and evaluation on the val2017 set (5k images) with 32 batch sizes using 8 V100 GPUs. Following the settings in \cite{chu2019fairnas}, we train the detection model with 12 epochs, an initial learning rate of 0.04, and multiply the learning rate by 0.1 at epochs 8 and 11. The optimizer is SGD with 0.9 momentum and 1e-4 weight decay. As shown in Table~\ref{tab:det_retina}, our method surpasses MobileNetV2 by 4.7\% using similar FLOPs. Compared with MnasNet \cite{tan2019mnasnet}, our method utilizes 7\% fewer FLOPs while achieving 2.5\% higher performance, suggesting the architecture has good generalization ability when transferred to other vision tasks.
	
\section{Conclusion}
In this work, we propose a novel approach to search for lightweight ensemble models based on one-shot NAS. We design a new metric, called \textit{diversity score}, to guide search space shrinking. We further use the layer-sharing mechanism to reduce the complexity of ensemble models and introduce a new search dimension, called \textit{split point}, to handle the trade-off between diversity and complexity constraint. Extensive experiments demonstrate that the proposed new metric is effective and improves the weight sharing supernet's ranking ability. Our searched architectures do achieve not only state-of-the-art performance on ImageNet but also have great generalization ability and robustness.  

{\small
\bibliographystyle{ieee_fullname}
\bibliography{egbib}
}
\clearpage
\onecolumn




\section*{Appendix A}
In this appendix, we include: (I) proof of the property stated in Section 3.2, (II) the detailed supernet structure and search space.

\vspace{2mm}\subsection*{A-I: Proof of Diversity Score Property}

In this section, we show a more detailed formula of the property stated in Section 3.2 and the proof of the property.

\noindent\textbf{Property:} Assume that $h_m:= (o_{1,m}, \cdots, o_{j,m},  \cdots ,o_{K,m})$ and $h_m^{\rm '} := (o_{1,m}, \cdots, o_{j,m}^{\rm '},  \cdots ,o_{K,m})$ are different only by $j_{th}$ operator. Denote the indexes of operators in $h_m$ and $h_m^{'}$ as $\sigma_1, \sigma_2, \cdots, \sigma_K$ and $\sigma_1^{\rm '}, \sigma_2^{\rm '}, \cdots, \sigma_K^{\rm '}$. If $S_{i,k}^m < S_{i^{\rm '},k}^m$ for $k = 1,2,\cdots,K$ and $r_{i}^m > r_{i^{\rm '}}^m$, then we have:
\begin{equation}
\begin{aligned}
&Score(h_m) > Score(h_m^{'}),
\end{aligned}
\end{equation}
where $\sigma_j$ and $\sigma_j^{'}$ equal to $i$ and $i^{'}$.

\noindent\textbf{Proof:} Given the property of matrix determinant and definition of $L_m^y$, the diversity score of $h_m$ could be expressed as:  
\begin{equation}
\begin{aligned}
Score(h_m) = \prod_{i=1}^{K} r_{\sigma_i}^2\cdot {\rm det}(S_m^y).
\end{aligned}
\label{eq:define}
\end{equation}
where $S_m^y$ are the corresponding submatrixs of $h_m$ in $S_m$.

According to the assumption, we know that $\prod_{i=1}^{K} r_{\sigma_i}^2 > \prod_{i=1}^{K} r_{\sigma_i^{'}}^2$. Now, if ${\rm det}(S_m^y)$ is greater than ${\rm det}(S_m^{y^{'}})$ then the property holds easily. Because $h_m$ and $h_m^{'}$ are only different by the $j_{th}$ operator and $S_m$ is a symmetry matrix, the number of total different entries between $S_m^y$ and  $S_m^{y^{'}}$ is less than $2K$. We could construct a series of matrixs $B_i \in \mathbb{R}^{K \times K}, i=0,1,2,\cdots, K$ as following:
\begin{equation}
B_i(k,l)=\left\{
\begin{aligned}
S_m^y(k,l),  &  & k<i,l=j, \\
S_m^y(k,l),  &  & l<i,k=j, \\
S_m^{y^{'}}(k,l), &  & Otherwise,\\
\end{aligned}
\right.
\end{equation}
where $B_i(k,l)$ is the entry in row $k$ column $l$.
We then prove the following inequality by induction:
\begin{equation}
{\rm det}(B_i) \leq {\rm det}(B_{i+1}), i=0,1,2,\cdots K-1.
\end{equation}
For $i=0$, consider matrix $A$ defined as follow:
\begin{equation}
A(k,l)=\left\{
\begin{aligned}
&\frac{S_m^y(1,j)}{S_m^{y^{'}}(1,j)},  &  & k=1,l=j, \\
&\frac{S_m^y(j,1)}{S_m^{y^{'}}(j,1)}, &  & l=1,k=j, \\
&\quad \ 1, &  & Otherwise.\\
\end{aligned}
\right.
\end{equation}
Given the assumption that $S_{i,k}^m < S_{i^{\rm '},k}^m$ for $k = 1,2,\cdots,K$, we have ${S_m^y(1,j)} < S_m^{y^{'}}(1,j)$. Then we could get $A$ is a positive define matrix easily using the definition of positive define matrixs. Regarding $A$ and $B_0$ are both semi-positive define matrix, we have following statement using \textbf{Oppenheim's inequality}:
\begin{equation}
{\rm det}(A\circ B_0) = {\rm det}(B_1) \ge {\rm det}(B_0)  \prod_{i}^K A(i,i) =  {\rm det}(B_0),
\end{equation}
where $A\circ B_0$ is the \textbf{Hadamard product} (element-wise product) of $A$ and $B_0$. Besides, $B_1$ is also a semi-positive define matrix according to \textbf{Schur product theorem}.

For $i=1,2,\cdots,K-1$, it is easy to construct $A$ with similar definition like above and get the statement that ${\rm det}(B_i)\leq {\rm det}(B_{i+1})$. Now, combining the chain of inequality, we have:
\begin{equation}
{\rm det}(S_m^y) = {\rm det}(B_{K-1}) \ge {\rm det}(B_0) =  {\rm det}(S_m^{y^{'}}).
\label{eq:similarity_new}
\end{equation}
Using Eq.~(\ref{eq:define})(\ref{eq:similarity_new}), the property holds easily.

\vspace{2mm}\subsection*{A-II: Supernet Structure and Search Space}
In this section we give the detailed supernet structer and space of the new dimension \textit{Splint Point}.
\begin{table}[h]
	\centering
	\begin{tabular}{ccccc}
		\toprule[0.75pt]
		\multirow{2}{*}{Input Shape} & \multirow{2}{*}{Operators} & \multirow{2}{*}{Channels} & \multirow{2}{*}{Repeat}  & \multirow{2}{*}{Stride} \\
		\\
		\midrule[0.75pt]
		$224^2\times3$  & $3\times3$ Conv    & 16    & 1 & 2\\
		$112^2\times16$  & $3\times3$ Depthwise Separable Conv    & 16    & 1 & 2\\
		$56^2\times16$ & MBConv / SkipConnect         & 24   & 4 & 2\\
		$28^2\times24$ & MBConv /  SkipConnect             & 40   & 4 & 2\\
		$14^2\times40$ & MBConv /  SkipConnect            & 80   & 4 & 1\\
		$14^2\times80$ & MBConv /  SkipConnect             & 112   & 4 & 2\\
		$7^2\times112$ & MBConv /  SkipConnect             & 160   & 4 & 1\\
		$7^2\times160$  & $1\times1$ Conv    & 960  & 1 & 1\\

		$7^2\times960$  & Global Avg. Pooling    & 960  & 1 & -\\
		$960$  & $1\times1$ Conv    & 1,280  & 1 & 1\\
		$1,280$ & Fully Connect & 1,000 & 1 & -\\
		\midrule[0.75pt]
		\midrule[0.75pt]
		\multicolumn{2}{c}{Split Point} & \multicolumn{3}{c}{(9, 20, 1)}\\
		\bottomrule[0.75pt]
		\vspace{0.3cm}
	\end{tabular}
	\caption{The structure of the supernet. The "MBConv" contains $6$ inverted bottleneck residual block MBConv \cite{sandler2018mobilenetv2} ( kernel sizes of \{3,5,7\}) with the squeeze and excitation module (expansion rates \{4,6\}). The "Repeat" represents the maximum number of repeated blocks in a group. The "Stride" indicates the convolutional stride of the first block in each repeated group. (9, 20, 1) means space starts from 9 to 20 with a step of 1.} 
	
	\label{tab:design}
\end{table}	

\section*{Appendix B}
In appendix B, we show the detailed evolution algorithm, with the detailed algorithm of $K$-path evolution search below. Specific steps of ${\rm Crossover},{\rm Mutation}$ are presented in Section 3.4.
\begin{algorithm}[h]
	\renewcommand{\algorithmicrequire}{\textbf{Input:}}
	\renewcommand{\algorithmicensure}{\textbf{Output:}}
	\caption{K-Path Evolution Search}
	\label{alg:EVO}
	\begin{algorithmic}[1]
		\REQUIRE \mbox{} \\
		 Shrunk search space $\tilde{\mathcal{S}}$, weights $W_{\tilde{\mathcal{S}}}$, population size $P$, resources constraints $C$, number of generation iteration $\mathcal{T}$, validation dataset $D_{\rm val}$, training dataset $D_{\rm train}$,  Mutation probability of split point $P_s$,  Mutation probability of layer combination $P_m$.
		\ENSURE The most promising ensemble architecture $\mathcal{A}^*$.
		\STATE $G_{(0)}:=$ Random sample $P$ ensemble architectures $\{\mathcal{A}_1, \mathcal{A}_2, \cdots \mathcal{A}_P\}$ from $\tilde{\mathcal{S}}$ with constrain $C$; 
		\WHILE {search step $t \in (0,\mathcal{T})$}
		\WHILE {${\mathcal{A}_i}\in G_{(t)}$}
		\STATE Recalculate the statistics of BN on $D_{\rm train}$;
		\STATE Obtain the accuracy of $\Phi(\cdot;\mathcal{A}_i,W_{\tilde{\mathcal{S}}})$ on $D_{\rm val}$.
		\ENDWHILE
		\STATE $G_{\rm topk}:=$ the Top $K$ candidates by accuracy order;
        \STATE $G_{\rm crossover}:= {\rm Crossover}(G_{\rm topk},\tilde{\mathcal{S}},C)$;
        \STATE $G_{\rm mutation}:= {\rm Mutation}(G_{\rm topk},P_s,P_m, \tilde{\mathcal{S}},C)$;
		\STATE $G_{(t+1)} = G_{\rm crossover} \cup G_{\rm mutation}$
		\ENDWHILE
	\end{algorithmic} 
\end{algorithm}

\end{document}